\documentclass{article}
\usepackage{ijcai25}

\usepackage{times}
\usepackage{soul}
\usepackage{url}
\usepackage[hidelinks]{hyperref}
\usepackage[utf8]{inputenc}
\usepackage[small]{caption}
\usepackage{graphicx}
\usepackage{amsmath}
\usepackage{amsthm}
\usepackage{booktabs}
\usepackage{algorithm}
\usepackage{algorithmic}
\usepackage[switch]{lineno}
\usepackage{cprotect}
\usepackage{footmisc}
\usepackage{fancyhdr}

\title{InkFM: A Foundational Model for Full-Page Online Handwritten Note Understanding}

\author{
Anastasiia Fadeeva \footnote{first author}
\and
Vincent Coriou\and
Diego Antognini\and
Claudiu Musat\and
Andrii Maksai\\
\affiliations
Google DeepMind\\
\emails
\{fadeich, vcoriou, dantognini, cmusat, amaksai\}@google.com
}

\begin{document}

\maketitle

\begin{abstract}
    Tablets and styluses are increasingly popular for taking notes. To optimize this experience and ensure a smooth and efficient workflow, it's important to develop methods for accurately interpreting and understanding the content of handwritten digital notes. We introduce a foundational model called \textbf{InkFM} for analyzing full pages of handwritten content. Trained on a diverse mixture of tasks, this model offers a unique combination of capabilities: recognizing text in 28 different scripts, mathematical expressions recognition, and segmenting pages into distinct elements like text and drawings. Our results demonstrate that these tasks can be effectively unified within a single model, achieving SoTA text line segmentation out-of-the-box quality surpassing public baselines like docTR. Fine- or LoRA-tuning our base model on public datasets further improves the quality of page segmentation, achieves state-of the art text recognition (DeepWriting, CASIA, SCUT, and Mathwriting datasets) and sketch classification (QuickDraw). This adaptability of InkFM provides a powerful starting point for developing applications with handwritten input.

\end{abstract}

\section{Introduction}
\label{sec:intro}

Large screens with precise tools like styluses are changing digital note-taking. They offer a large canvas for note takers, similar to a large context window for Large Language Models (LLMs) - where users can interact with all the elements that are needed in their current context. This creates exciting possibilities for better note-taking tools that can understand and generate handwritten content, easing the note takers' access to the concepts within the note. 

Given this large-form input, tools can surface advanced assistance capabilities to the user, beyond the basic transcription of the text - they can identify what is drawn or written, by segmenting the different parts of a note and classifying what each part represents. For example, they can distinguish between a diagram, a table, or a paragraph of text \cite{10.1145/3613904.3642498}. They can also convert handwriting to typed text, a task known as handwriting recognition \cite{fadeeva2024representingonlinehandwritingrecognition}. They can fix mistakes using error correction algorithms and can even make writing look neater with handwriting beautification features \cite{maksai2022inkorrectonlinehandwritingspelling}. 

\paragraph {Note-taking tasks} Useful note taking tools are thus complex, merging both understanding and synthesis features. All these features traditionally require disparate models to function \cite{10.1145/3613904.3642498}. This paper focuses on three key tasks essential for understanding notes, which are usually done by different models. We introduce a new way to combine them into one powerful LLM. These tasks are:
\begin{enumerate}
    \item Segmentation: assigning strokes to objects, classifying each instance into separate objects belonging to different classes, for instance text or drawings. For example, on the IAM dataset \cite{IAMonDB}, we focus on line segmentation.
    \item Classification: figuring out what an object is, based on its strokes – is it a drawing, a table, or text? For example, on the QuickDraw dataset \cite{DBLP:journals/corr/HaE17} we classify sketches into categories like "squirrel" or "picture frame".
    \item Recognition: transcribing the strokes assigned to textual objects into sequences of characters. 
\end{enumerate}

\paragraph{Why use large models?} Small models can be limited and brittle. The intuition behind merging multiple tasks into a single LLM is that the model will be able to transfer knowledge from one task to another, while not observing significant negative transfers. In note taking for instance, we can envision classification knowledge being used to improve the separation of stroke-based objects, and classification to improve character, and broadly textual, recognition.
Moreover, using a single LLM uses its intrinsic representation capability and has significant distribution and model maintenance advantages.
Conversely, adding these inking tasks to a model can improve its localization and vision capabilities outside of the inking domains. Furthermore, future extensibility is a major advantage of a unified model, with generative tasks like ink synthesis being possible for additional user assistance. 

\paragraph{Scope of this work} We define a mixture of tasks in the ink domain. We then use it to fine-tune PaliGemma, due to its demonstrated adaptability across a variety of tasks \cite{beyer2024paligemmaversatile3bvlm}. 
Our results demonstrate that this unified approach is effective across a variety of handwriting styles and tasks. In the segmentation of handwritten text lines, we achieve strong performance on the English language IAM dataset. When classifying sketches, our model shows state of the art accuracy on the QuickDraw dataset. Furthermore, our model is able to recognize handwritten text effectively across 28 scripts. For instance, we achieve competitive results on datasets like IAM (English), CASIA \cite{liu2011casia}, SCUT \cite{SCUT-COUCH2009}, or MathWriting \cite{gervais2024mathwritingdatasethandwrittenmathematical}, demonstrating the model's adaptability to different writing systems.
We further show that parameter efficient fine-tuning (PEFT) - in this case LoRA \cite{hu2021loralowrankadaptationlarge} on top of the foundational model further improves segmentation and certain recognition results. This work makes the following key contributions: 
\begin{itemize}
\item We propose a method that unifies segmentation, recognition and classification of handwriting in one model; we show the model's extensibility with parameter efficient fine tuning. We plan to make this model public.
\item We show the SoTA quality on handwritten text line segmentation (English), sketch classification and multi-script text recognition on 5 out of 6 datasets.
\item Our model demonstrates production-grade performance in detecting visual elements within English handwritten notes, including lists, tables, and diagrams.
\end{itemize}
\section{Related work}
\label{sec:related_work}
The development of foundational models has become a significant area of research in recent years. 
The idea is to leverage a model pre-trained on a massive dataset – \textbf{foundational model} and then use it as a starting point for the tasks of interest \cite{brown2020languagemodelsfewshotlearners}. Complex tasks like code generation \cite{huang2024agentcodermultiagentbasedcodegeneration} and math reasoning \cite{hendrycksmath2021} are currently solved by using Large Language Models like GPT-4  \cite{openai2024gpt4technicalreport} and Gemini-2  \cite{geminiteam2024geminifamilyhighlycapable}. Many recent foundational models like CLIP \cite{radford2021learningtransferablevisualmodels}, Flamingo \cite{alayrac2022flamingovisuallanguagemodel}, PaLI \cite{chen2023palijointlyscaledmultilinguallanguageimage} and GPT-4 are \textbf{multimodal}, accepting both text and images as input. This enables capabilities like Visual Question Answering \cite{goyal2017makingvvqamatter} and Image Captioning \cite{lin2015microsoftcococommonobjects} in foundational models.

The task of \textbf{Optical Character Recognition} (OCR) aims to accurately recognize text within complex documents. This technology has various applications, including digitizing historical texts, automating data entry for forms and invoices, and facilitating document translation. Traditionally, approaches to OCR relied on models trained from scratch like RNN encoders with CTC decoders \cite{8270042} or LSTM encoder-decoder models \cite{Michael2019EvaluatingSM}. The success of transformer architectures in natural language processing led to their adoption for OCR tasks, with significant improvements seen in \cite{diaz2021rethinkingtextlinerecognition}. Image pre-training has been shown to significantly benefit OCR tasks by enabling models to ignore background noise, handle low-resolution images, and adapt to various fonts \cite{li2022trocrtransformerbasedopticalcharacter}. Furthermore, language information is also aiding in the accurate decoding of written text \cite{10483819}. This encourages wider adoption of pre-trained multimodal models for OCR tasks.

Unlike OCR, which primarily deals with static images of text, \textbf{online handwriting domain} leverages both spatial and temporal information. This means it analyzes not just the shape of the written characters, but also the order and speed of strokes. Online handwriting recognition, especially for languages like Chinese with complex characters, has been a popular focus of attention in AI research \cite{9932793}. The online handwriting domain encompasses a wide range of tasks beyond just recognition, including complex challenges like understanding handwritten notes \cite{10.1145/1815330.1815343}. These tasks are leveraged to make writing on a tablet feel more natural and intuitive. Although there have been advancements in online handwriting recognition using deep learning models, a comprehensive foundational model that generalizes across different tasks and languages is still lacking. \textbf{Our work} To the best of our knowledge we are the first to utilize pre-trained multimodal foundational model for full-page handwritten note understanding.

\section{Method}
\label{sec:method}
Our work focuses on developing one foundational model \textbf{InkFM} for understanding online handwriting, enabling full-page segmentation, recognition and sketch classification.

\begin{figure}[!ht]
    \centering
    \includegraphics[width=200px]{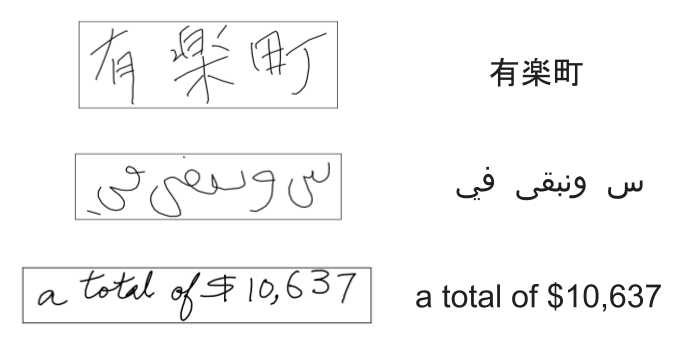}
    \caption{Examples of handwriting in Japanese, Arabic, and English.}
    \label{fig:multi_script}
\end{figure}

As a foundation we use the open-source multimodal model \textbf{PaLIGemma} \cite{beyer2024paligemmaversatile3bvlm} which consists of SigLIP \cite{zhai2023sigmoidlosslanguageimage} 400M vision encoder and Gemma \cite{gemmateam2024gemmaopenmodelsbased} 2B decoder. PaLIGemma architecture is similar to the PaLM-E model which was used in \cite{fadeeva2024representingonlinehandwritingrecognition} for handwriting recognition. 
The inclusion of OCR task in PaLIGemma's pre-training mixture is a key advantage, as it equips the model with the ability to effectively analyze printed text and, with further fine-tuning, adapt to handwritten inputs (example Figure~\ref{fig:multi_script}).

We fine-tune PaLIGemma-3B on a mixture of handwriting tasks like multi-script text and math recognition, classification and segmentation. This training enables the creation of a single, versatile model capable of tackling a variety of handwriting tasks. 

\subsection{Mixture of Handwriting Tasks}
The goal is to develop a diverse dataset of handwriting tasks that enables a single model to learn a wide range of skills in handwriting domain.

\subsubsection{Note Segmentation}
\label{sec:method_note_segmentation}

\begin{figure}[!ht]
    \centering
    \includegraphics[width=250px]{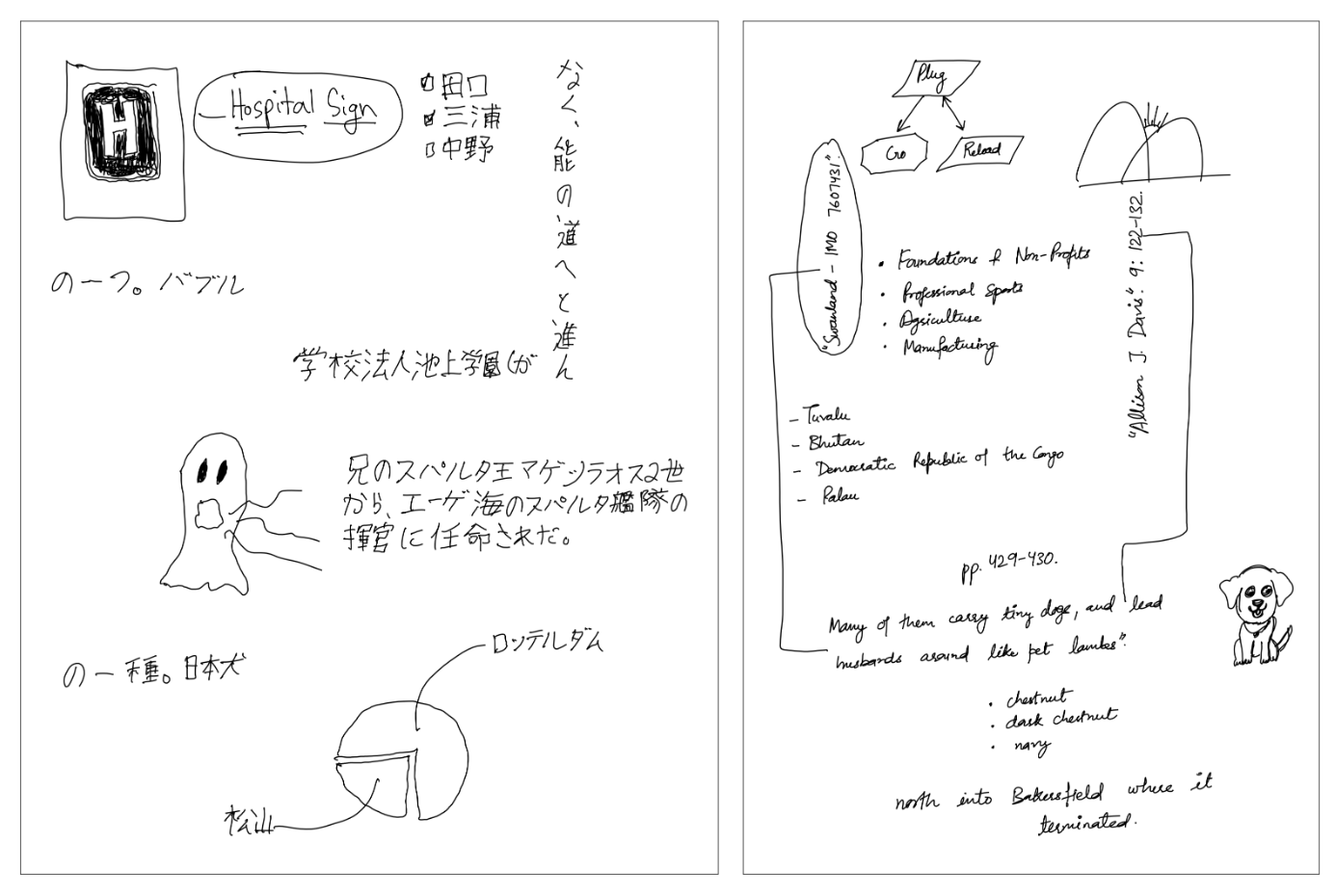}
    \caption{Examples of notes in Japanese (left) and English (right).}
    \label{fig:fullpage_example}
\end{figure}

Handwritten note segmentation is a crucial skill for models, enabling them to accurately distinguish between tables, lists, drawing and text within a page.
This functionality enables precise object selection in note-taking apps, improving the user experience \cite{Delaye2014ContextualTS,VANPHAN2016112}.

We use a private dataset that consists of 70\% English pages, 10\% Japanese, 10\% Devanagari and 10\% Telugu, examples are shown in Figure~\ref{fig:fullpage_example}. The dataset consists of 22k unique full-page handwritten notes, each segmented into three hierarchical levels. The first level captures the most granular elements, such as individual words, while the third level represents the broadest units, like complete text blocks, see Table~\ref{tab:segmentation_tasks_table}.
\begin{table}[!ht]
\caption{Segmentation classes statistics.}
\small
  \centering
\begin{tabular}{lccc}
\toprule
level & class & \% present & avg \#  \\ 
\midrule
2 & textblock & 94.61 & 4.58 \\
2 & diagram & 53.05 & 0.53 \\
2 & list & 76.87 & 1.19 \\
2 & drawing & 65.7 & 0.98 \\
2 & table & 6.34 & 0.06 \\
\midrule
1 & textline & 99.98 & 15.5 \\
1 & enclosure & 51.09 & 0.75 \\ 
\midrule
0 & word & 99.98 & 49.54 \\ 
0 & arrow & 74.15 & 2.08 \\
0 & oval & 41.27 & 0.58 \\
0 & box & 34.32 & 0.46 \\

\bottomrule
\end{tabular}
\label{tab:segmentation_tasks_table}
\end{table}

\begin{figure}[!ht]
    \centering
    \includegraphics[width=250px]{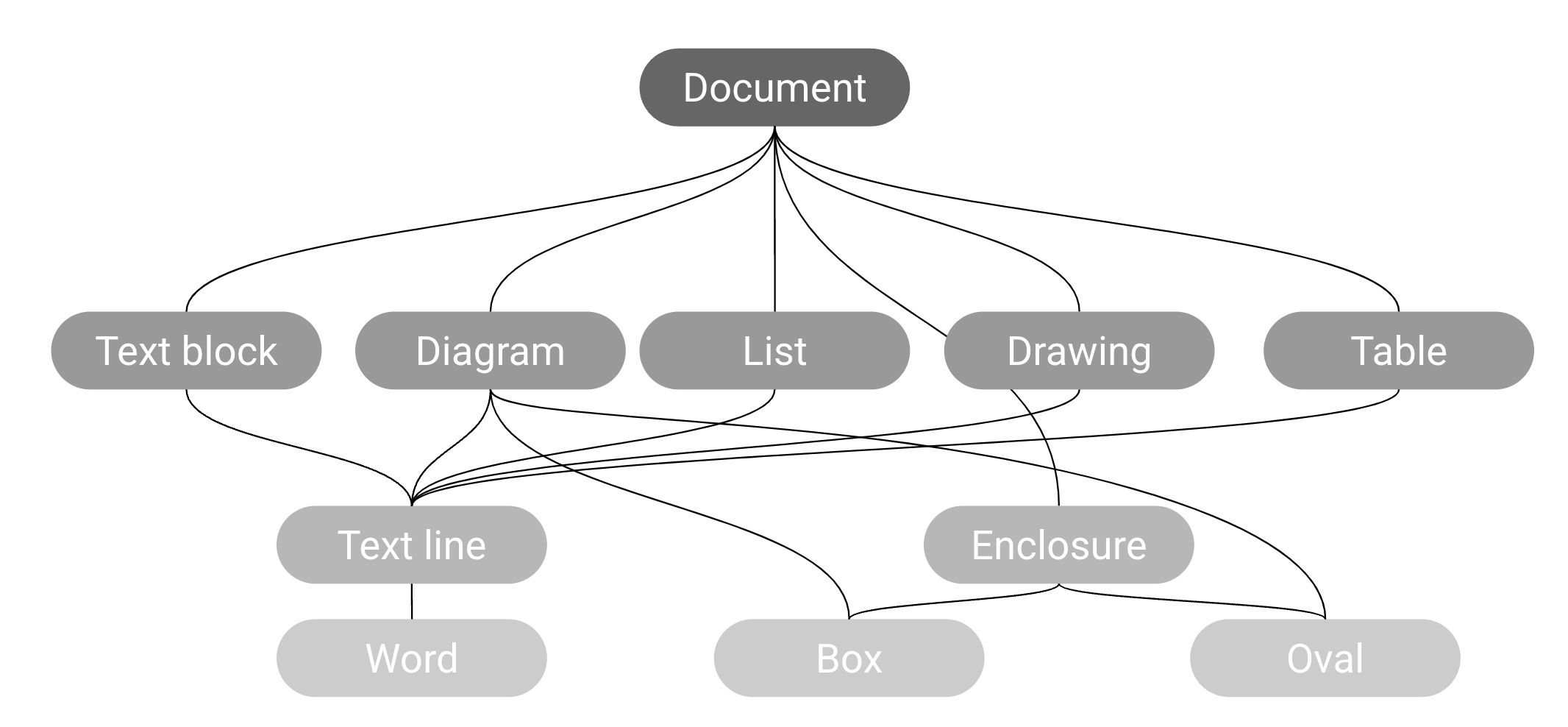}
    \caption{Three levels of segmentation in a full-page handwritten note.}
    \label{fig:levels_scheme}
\end{figure}

We define two types of tasks for segmentation. In the first type, we ask the model to identify all instances of \textbf{one type of object}.
\begin{itemize}
    \item \textbf{Question} Where are the level 1 objects located? Detect multiple textline
    \item \textbf{Target} ``38 41 67 273 textline 94 106 118 200 textline ..."
\end{itemize}

In the second type, we extend the task 1 to \textbf{mutiple} object types
\begin{itemize}
    \item \textbf{Question} Where are the level 1 objects located? Detect multiple enclosure, textline
    \item \textbf{Target} ``278 243 296 269 enclosure 1 114 11 303 textline 15 113 29 304 textline ..."
\end{itemize}
 
We found that it is better to list all objects of one class together in the target and match the order of classes to the prompt. So, if in the prompt we listed classes $c_1, c_2, c_3$, then multi-class target will have the following form:
\begin{equation}
    \text{join}(\text{target}({c_1}), \text{target}(c_2), \text{target}(c_3))
\end{equation}
where $\text{target}$ is single-class target provided in the first task type. Our findings in Section~\ref{sec:segmentation_ablation} demonstrate the value of both task formulations.  Therefore, we utilize an equal mix of each during the training.



\subsubsection{Multi-script Text Recognition}
\label{sec:multi_script}

Online handwriting recognition is a popular and important task due to its applications in areas like note-taking. It is important to note that handwriting in different languages presents a unique set of challenges. For instance, writing systems like those used for Chinese, Japanese, and Korean have thousands of logographic characters and have been a focus of specialised research \cite{Altwaijry2020ArabicHR,Zhang2016DrawingAR}. Traditionally, recognition models were trained on a single script, limiting their ability to handle a mix of scripts in one line \cite{carbune2020recognition}.

In order to combine recognition on multiple scripts, it's essential to utilize an appropriate \textbf{online ink representation} that captures the nuances of each script. We utilize findings from \cite{fadeeva2024representingonlinehandwritingrecognition} where inks were rendered with time and distance information in the color channels. Let's denote an ink $I$ as a sequence of strokes $s_i$  and each stroke is a sequence of coordinates with time information $s_i = [(x_{i, 0}, y_{i, 0}, t_{i, 0}), \ldots, (x_{i, n}, y_{i, n}, t_{i, n})]$
where $n$ is a number of points in $I$. Then, we can render ink $I$ with time and distance information as follows:
\begin{align}\label{eq:speed_rendering}
c^{R}_{i, j} = \frac{t_{i, j} - t_{0, 0}}{\max t_{m, n}} ~~
c^{G}_{i, j} = \frac{|d x_{i, j}|}{\max |d x_{m, n}|} ~~
c^{B}_{i, j} = \frac{|d y_{i, j}|}{\max |d y_{m, n}|}
\end{align}

\begin{figure}[!ht]
    \centering
    \includegraphics[width=150px]{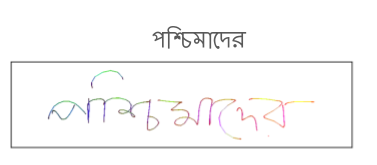}
    \caption{Example of Bengali writing with time and distance rendering.}
    \label{fig:speed_rendering}
\end{figure}

The inks are fitted to the image size preserving the aspect ratio, see Figure \ref{fig:speed_rendering}. This representation has been shown to be effective when used with VLMs \cite{fadeeva2024representingonlinehandwritingrecognition}.

Together with an image of handwriting we provide a \textbf{textual prompt} of the task. The prompt contains these key elements:
\begin{itemize}
    \item \textbf{Question} What is written in the image?
    \item \textbf{Language} English
    \item \textbf{Precontext} to be or not to
    \item \textbf{Placement} 0.39,0.15,0.67,0.78
\end{itemize}

\noindent \fbox{\begin{minipage}{\linewidth}
What is written in the image? Language \{\} Precontext \{\} Placement \{\}
\end{minipage}}

Following the work on multitask prompted training, distinct question prompts are used to guide the model towards the appropriate task within a mixture of tasks.
Similarly, information about the language makes it easier to do recognition. Our ablation study (Table~\ref{table:language_code_abaltion}) shows that providing language information significantly improves recognition quality for less common languages like Vietnamese. Precontext information consists of the words that appeared before the current point in writing and allows to use the language modeling capabilities of the VLM. A non-empty precontext is present in less than 1\% of the training data.  The placement information is a relative position of an ink bounding box within the writing area, and it helps distinguish between potentially ambiguous characters that can only be differentiated by their relative location, such as the comma and the apostrophe. The formula for the placement is provided in Equation~\ref{eq:placement} where $w$ is width and $h$ is height of the writing canvas.
\begin{align}\label{eq:placement}
\left[ \frac{x_{min}}{w}, ~~
\frac{y_{min}}{h},
\frac{x_{max}}{w}, ~~
\frac{y_{max}}{h} \right]
\end{align}

\begin{figure}[!ht]
    \centering
    \includegraphics[width=250px]{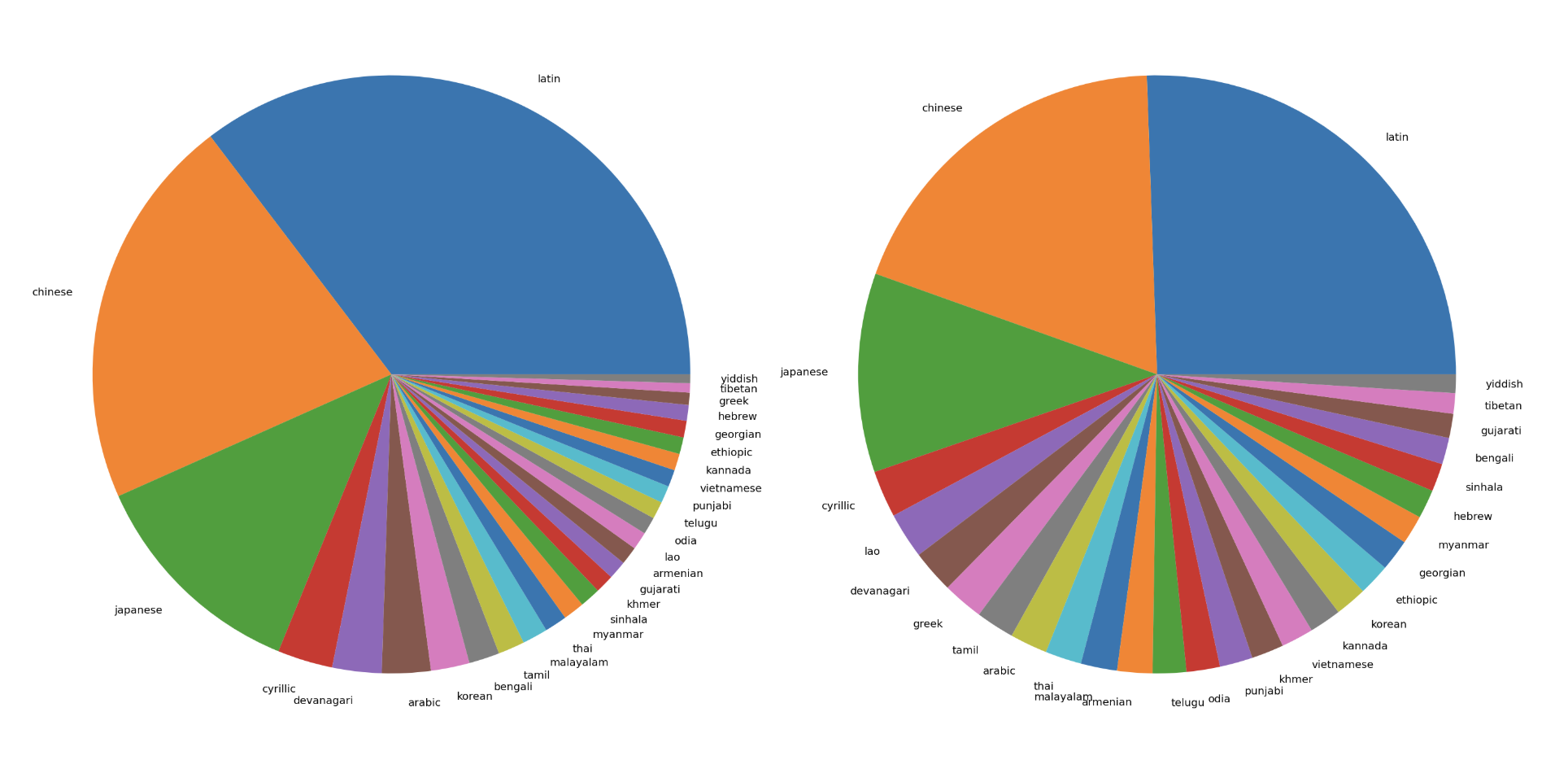}
    \caption{\textbf{Left:} Original distribution of different languages. \textbf{Right:} Adjusted distribution.}
    \label{fig:script_distribution}
\end{figure}

The training dataset is private and contains 41 million samples and Latin languages make up 35\% of it. The next biggest language is Chinese which covers 21\%. 
Figure \ref{fig:script_distribution} illustrates the complete distribution across languages. This presents a challenge in creating a mixture that effectively represents both common and rare languages. To address the challenges of imbalance in the dataset we adjusted the mixing weights to ensure that even the least frequent languages have a minimum share of 0.01. This technique was inspired by recent research on multilingual training for machine translation \cite{chen2023paretomultilingualneuralmachine}.

\subsubsection{Math Recognition}
\label{section:math_method}

\begin{figure}[!ht]
    \centering
    \includegraphics[width=150px]{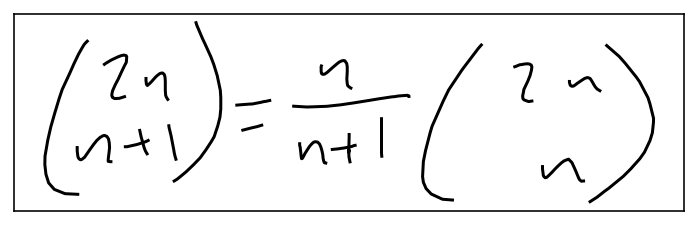}
    \cprotect\caption{Label: $\left(\begin{matrix}2n\\ n+1\end{matrix}\right)=\frac{n}{n+1}\left(\begin{matrix}2n\\ n\end{matrix}\right)$}
    \label{fig:math_example}
\end{figure}

Recognition of handwritten mathematical expressions is crucial for understanding documents that contain scientific and educational information \cite{10.1145/1815330.1815343}. Unlike multi-script recognition, mathematical expressions exhibit a complex two-dimensional structure. It was demonstrated by \cite{TRUONG2024110531} that common architectures like encoder-decoders can achieve high recognition quality. Leveraging that insight, we use the same image representation for an online ink as in multi-script recognition and adapt the textual prompt in the following way:
\begin{itemize}
    \item \textbf{Question} What is written in the image in LaTeX?
    \item \textbf{Language} LaTeX
    \item \textbf{Precontext} -
    \item \textbf{Placement} 0.39,0.15,0.67,0.78
\end{itemize}

For the training mixture we use a public dataset MathWriting \cite{gervais2024mathwritingdatasethandwrittenmathematical} that contains 230k human-written samples as well as 400k synthetic ones. We use normalized LaTeX label as a target as it was shown in \cite{gervais2024mathwritingdatasethandwrittenmathematical} to improve recognition quality.

\subsubsection{Classification}
\label{section:method_classification}
\begin{figure}[!ht]
    \centering
    \includegraphics[width=200px]{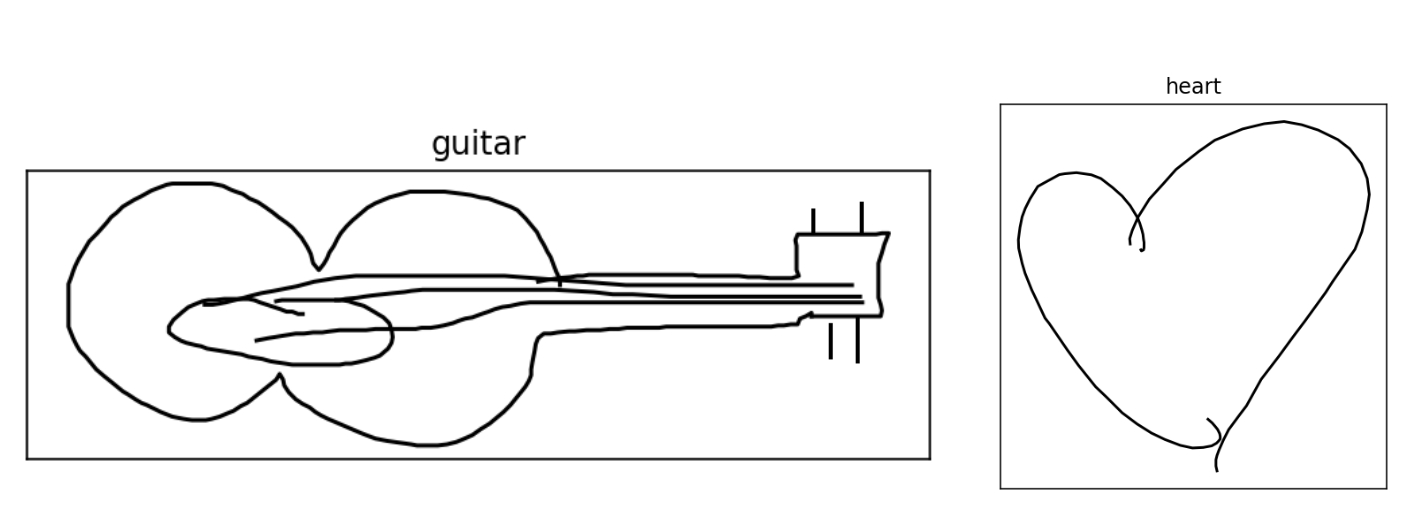}
    \caption{Example of a sketch from the QuickDraw dataset.}
    \label{fig:classification_example}
\end{figure}

Handwritten notes often contain illustrative sketches, so classifying them is a natural extension of our training mixture. We use a subset of a public dataset QuickDraw \cite{DBLP:journals/corr/HaE17} and a smaller private dataset with shape classification (see Figure~\ref{fig:classification_example}). We also incorporate a language detection task using the multi-script text dataset described previously. For this task we sample $\sim 50k$ samples per script as training data. Different prompts help the model to distinguish between the tasks:
\begin{itemize}
    \item \textbf{Question for sketches} What is drawn in this sketch?
    \item \textbf{Question for language detection} What script is this text written in?
\end{itemize}

\begin{table}[!ht]
\caption{Classification task statistics.}
\small
  \centering
\begin{tabular}{lccc}
\toprule
task & \# samples & \# classes & weight  \\ 
\midrule
QuickDraw & 450k & 345 & 0.48 \\ 
Shape & 40k & 194 &  0.11 \\ 
Languages & 1.3M & 28 & 0.41 \\ 
\bottomrule
\end{tabular}
\label{tab:classification_tasks_table}
\end{table}

The number of samples available for each classification task differs substantially. This can lead to biased model performance, favoring tasks with larger datasets. The number of classes to be predicted also varies across tasks. This introduces an inherent difference in task complexity, as tasks with more classes are generally more challenging. In order to evaluate the difficulty of each task we train models on tasks separately and calculate weights in Table~\ref{tab:classification_tasks_table} based on the number of steps needed to achieve plateau on validation accuracy.
\section{Experiments}
\label{sec:results}

\begin{table}[!ht]
\caption{Training task statistics for InkFM.}
\small
  \centering
\begin{tabular}{lccc}
\toprule
task & \# samples & seq length & weight \\ 
\midrule
Note segmentation & 2.4M & 512 & 15\% \\
Multi-script recognition & 40.8M & 128 & 50\% \\
Math expression recognition & 630k & 128 & 15\% \\
Classification & 1.8M & 32 & 20\% \\
\bottomrule
\end{tabular}
\label{tab:dataset_stats}
\end{table}
\subsection{Model}
We leverage the PaLIGemma-3B model \cite{beyer2024paligemmaversatile3bvlm} due to its demonstrated adaptability across various tasks. We use the checkpoint with 448px resolution in order to accurately segment smaller objects like arrows in the notes. We use the sequence length of 512 to cover 90\% of note segmentation targets. The inference measurements for refilling 512 tokens are provided in Appendix I of \cite{beyer2024paligemmaversatile3bvlm}.

\subsection{Training}

In the training dataset we mix the tasks from Section~\ref{sec:method} with the weights presented in Table~\ref{tab:dataset_stats}. We've picked the weights based on the needed number of steps for each task to achieve the plateau on validation. We fine-tune PaLIGemma for 5 epochs with 1024 batch size and learning rate 1e-5 (200M seen examples, 200k training steps, 18k TPUv5 core hours). The training took 3 days on 256 TPUv5e.
We call the PaLIGemma-3B model trained on a mixture of handwriting tasks – \textbf{InkFM}.

\subsection{Note segmentation results}
\begin{figure*}[!ht]
    \centering
    \includegraphics[width=\textwidth]{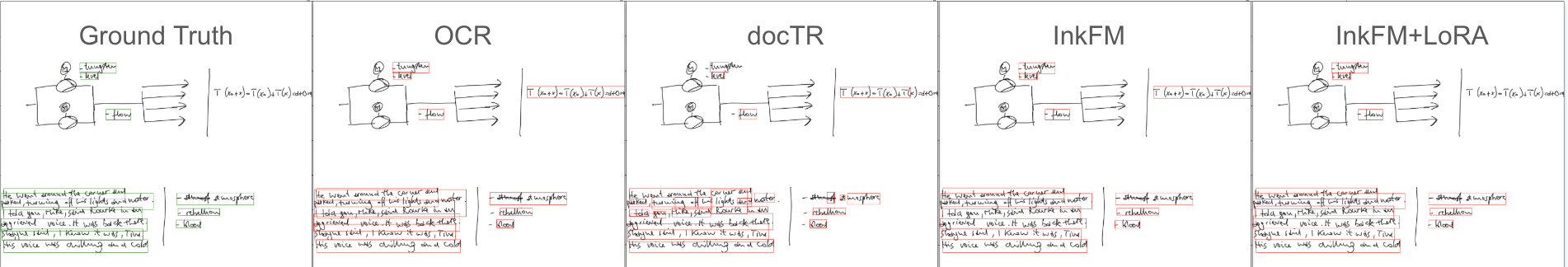}
    \caption{Text line segmentation results on an example from IAMonDo dataset.}
    \label{fig:segmentation_comparison}
\end{figure*}

To evaluate the quality of note segmentation, we use the IAMonDo dataset \cite{10.1145/1815330.1815343}, which consists of English full-page notes, see Figure~\ref{fig:segmentation_comparison}. We use the primary detection metric mean Average Precision from standard COCO metrics python package \cite{lin2015microsoftcococommonobjects} to evaluate the performance of the model as well as mean Average Precision at 50\% Intersection over Union. We compare our method to 
different detection models from popular public OCR models. The docTR model \cite{doctr2021} is open-source and it uses DBNet~\cite{db_res} with ResNet-50 backbone (25M parameters) as segmentation model. We also evaluate Google Cloud Vision segmentation results which has support support for handwriting and is considered a good public solution \cite{OCR_review}. docTR and Google Cloud Vision segmentation were evaluated (zero-shot) with 896px resolution to ensure optimal performance.

\begin{table}[!ht]
  \caption{
  Comparing \textbf{textline} segmentation capabilities of the mixture model with baselines.
 }
  \label{table:seg_result}
  \small
  \centering
  \begin{tabular}{lccc}
  \toprule
    & \multicolumn{2}{c}{IAMonDo } \\
  Model  & mAP $\uparrow$ & mAP@50IoU $\uparrow$ \\
  \midrule
   docTR~\cite{doctr2021} & 20.00 & 30.70 \\
   OCR~\cite{GoogleOcr} & 36.82 & 52.89 \\
  \midrule
   \textbf{InkFM [ours]} & 44.11 & 59.34 \\
   \textbf{+ LoRA [ours]} & \textbf{51.05} & \textbf{66.15}  \\
    \bottomrule
  \end{tabular}
\end{table}

\begin{table}[!ht]
  \caption{The segmentation performance of a foundational model on a variety of classes.}
  \label{table:seg_classes}
  \small
  \centering
  \begin{tabular}{lcccc}
  \toprule
    & \multicolumn{2}{c}{InkFM} & \multicolumn{2}{c}{+LoRA} \\
  class  & mAP $\uparrow$ & mAP@50IoU $\uparrow$ & mAP $\uparrow$ & mAP@50IoU $\uparrow$ \\
  \midrule
   textblock & 12.59  & 19.34 & 38.08 & 48.74 \\
   list & 71.01  & 79.22 & 81.74 & 89.53 \\
   table & 72.31  & 86.84 & 88.60 & 94.52 \\
   diagram & 38.95 & 52.76 & 69.21 & 83.58\\
   \midrule
   textline & 44.11 & 59.34 & 51.05 & 66.15 \\
   enclosure & 74.87 & 86.54 & 75.56 & 86.45 \\
    \bottomrule
  \end{tabular}
\end{table}

\begin{table*}[!ht]
  \caption{Comparison of the foundational model on the multi-script recognition. Methods like DTrOCR, PaLI and PaLM-E also use pre-trained checkpoints. Proposed model with fine-tuning achieves the best results on 5 out of 6 datasets.
  }
  \label{table:reco_result}
  \small
  \centering
  \begin{tabular}{lcccccc}
  \toprule
    & \multicolumn{6}{c}{CER $\downarrow$} \\
  Model  & IAMonDB & DeepWriting & VNOnDB & MathWriting & CASIA & SCUT\\
    \midrule
    MyScript~\cite{nguyen2018icfhr} & - &  - & 2.91 & - & - & -\\
    LSTM+CTC~\cite{carbune2020recognition} &  2.5 & 
     6.14 & 4.13 & - & 4.69 & 0.60\\
    OCR~\cite{GoogleOcr}  & 4.32 & 14.16 & 4.83 & 5.93 & 13.18 & 1.99\\
    DTrOCR~\cite{10483819} & \textbf{2.38} & - & - & - & - & - \\
     

     PaLI \cite{fadeeva2024representingonlinehandwritingrecognition} & - & 4.09  & 2.81 & 4.27 & 3.89 & 0.74\\
     
     PaLM-E \cite{fadeeva2024representingonlinehandwritingrecognition} & - & 6.53  & 3.05  & 4.05  & 4.27 & 0.87   \\
     
    \midrule
    \textbf{InkFM [ours]} & 3.52 &  6.47 & 3.66 & 4.53 & 3.99 & 0.411  \\
    
    \textbf{InkFM + LoRA [ours]} & 3.01 &  \textbf{3.90} & \textbf{2.55} & -  & -  & -  \\ 
    \textbf{InkFM + SFT [ours]} & -  & - & -  & \textbf{3.88} & \textbf{3.35} & \textbf{0.350}  \\
    \bottomrule
  \end{tabular}
\end{table*}

Our foundational model demonstrates better out-of-the-box performance in text line segmentation compared to public benchmarks, see Table~\ref{table:seg_result}.  Training LoRA \cite{hu2021loralowrankadaptationlarge} on top of the foundational model further improves these results, achieving even greater mean average precision. We've trained LoRA with rank 16 with learning rate 5e-4 for 10 epochs (the learning rate was picked out of 4 options based on minieval performance). The training of 1.2\% of additional weights took less than 30 minutes on 64 TPUv5e. The model is trained to detect all classes within the IAMonDo dataset. Further evaluation details are provided in the Table~\ref{table:seg_classes}.

A visual assessment of the text line segmentation in Figure~\ref{fig:segmentation_comparison} shows that the docTR method sometimes misses portions of words, leading to incomplete coverage. Additionally, the OCR struggles with the densely packed text in the left corner. Whereas, InkFM model performs consistently good across the example. 
LoRA adaptation eliminates the math formula in the top right corner of the note as math formulas are considered a separate class in IAMonDo dataset, improving overall performance.

The Figure~\ref{fig:segmentation_comparison} demonstrates the diverse nature of handwritten notes, which often include elements like diagrams, bullet lists, and mathematical formulas. The InkFM model is able to detect a diverse set of objects as shown in Table~\ref{table:seg_classes}. We see the strong out-of-the-box performance of our foundational model on lists, tables and enclosures. The lower evaluation performance on text blocks, diagrams, and text lines may stem from inconsistencies in annotation guidelines, for instance the inclusion of elements like list bullets within text lines (see Figure~\ref{fig:segmentation_comparison}). This ambiguity in annotation criteria likely contributes to discrepancies between the model's output and the ground truth labels, impacting the overall evaluation scores. Therefore, LoRA fine-tuning leads to significant improvements on those classes, as demonstrated by our results.

\subsection{Recognition Results}
\label{sec:recognition_results}

To evaluate the quality of our multi-script recognition, we use a variety of public datasets. IAMonDB \cite{IAMonDB} is a smaller dataset with 5k examples in English, while DeepWriting \cite{aksan2018deepwriting} is larger, containing 34k examples. We also evaluate on two datasets in Chinese – CASIA \cite{liu2011casia} and SCUT \cite{SCUT-COUCH2009} that contain 2.9M and 916k training examples respectively. To assess performance on under-represented languages, we evaluated the model on the public Vietnamese dataset VNOnDB \cite{nguyen2018icfhr}. Vietnamese examples constituted only 1.7\% of the recognition mixture (see Figure~\ref{fig:script_distribution}). VNOnDB dataset itself contains 34k training examples. We also report the results of mathematical expression recognition on MathWriting dataset that was used in training, see Section~\ref{section:math_method}.

We use the character error rate (CER) as our evaluation metric to compare the proposed foundational model to the state-of-the-art OCR (OCR~\cite{GoogleOcr}, DTrOCR~\cite{10483819}) and online handwriting recognition models (LSTM+CTC~\cite{carbune2020recognition}, MyScript~\cite{nguyen2018icfhr}, as well as results of fine-tuning PaLI (700M) and PaLM-E (500M) \cite{fadeeva2024representingonlinehandwritingrecognition}).

In Table ~\ref{table:reco_result} we show that the proposed model performs similarly or better, in case of the SCUT dataset, to previous approaches. We further improve the results by LoRA tuning on smaller datasets, like IAMonDB, DeepWriting and VNOnDB, and fine-tuning on bigger datasets, like CASIA, SCUT and MathWriting. This choice was selected based on the results from Section~\ref{sec:recognition_ablation}. We've trained LoRA with rank 16 with learning rate 1e-3 and batch size 128 for 10 epochs. The training took between 1h and 8h on 16 TPUv5e. For SFT we used learning rate 5e-5 for Chinese datasets and 1e-4 for MathWriting. We've trained with batch size 512 for 5 epochs and it took between 3h and 1d on 64 TPUv5e.

\subsection{Classification Results}
\begin{table*}
  \caption{Our foundational model InkFM performs better than LoRA and similar to SFT on a range of recognition tasks.}
  \label{table:reco_ablation}
  \small
  \centering
  \begin{tabular}{lcccccc}
  \toprule
    & \multicolumn{6}{c}{CER $\downarrow$} \\
  Model  & IAMonDB & MathWriting & VNOnDB & DeepWriting & CASIA & SCUT\\
    \midrule
    InkFM & \textbf{3.52} & 4.53 & \textbf{3.66} & 6.47  & 3.99 & 0.411  \\
    PaLIGemma SFT & 3.66 & \textbf{3.91} & 4.46 & 9.31 &  \textbf{3.50} & \textbf{0.269}\\
    PaLIGemma LoRA & 5.92 & 5.90 & 4.09 & \textbf{6.20} & 4.24 & 1.35\\
    \bottomrule
  \end{tabular}
\end{table*}
\begin{table*}[!h]
  \caption{Providing language information has a negligible effect on the recognition quality of high-resource languages such as Chinese and English. However, it is extremely important for less common languages like Vietnamese.}
  \label{table:language_code_abaltion}
  \small
  \centering
  \begin{tabular}{lccccc}
  \toprule
    & \multicolumn{5}{c}{CER $\downarrow$} \\
  Language  & IAMonDB & DeepWriting & VNOnDB  & CASIA & SCUT\\
    \midrule
    yes & 3.52 & 6.47 & \textbf{3.66}  & 3.99 & 0.411 \\
    no & 3.60 & 6.35 & 8.23 & 3.99 & 0.412 \\
    \bottomrule
  \end{tabular}
\end{table*}

\begin{table}[!ht]
  \caption{
  Comparison QuickDraw classification accuracy of the foundational model with BERT and other baselines.
 }
  \label{table:classification_result}
  \small
  \centering
  \begin{tabular}{lc}
  \toprule
  Model  & accuracy $\uparrow$ \\
  \midrule
  Transformer & 85.55 \\
  ResNet50~\cite{7780459} & 86.03 \\
  Sketch-BERT~\cite{Sketch_BERT} & 88.30 \\
  \midrule
   \textbf{InkFM [ours]} & 86.58 \\
   \textbf{+ SFT [ours]} & \textbf{91.00} \\
    \bottomrule
  \end{tabular}
\end{table}

To evaluate the quality of classification, we use the QuickDraw dataset \cite{DBLP:journals/corr/HaE17} that was partially used in the training, see Section~\ref{section:method_classification}. The full public dataset contains 50M samples. We compare the proposed model to fine-tuned Sketch-BERT~\cite{Sketch_BERT}  which represents sketches as sequences of strokes and utilizes a pre-trained checkpoint. We also train a transformer classifier from scratch with 8 layers, 256 layer size, 8 attention heads, drop-out 0.1 and 5e-4 learning rate. Ink is represented as a sequence of points similar to Sketch-BERT. We use this model as a baseline that doesn't rely on a pre-trained base. The proposed model out-of-the-box achieves better accuracy than ResNet50~\cite{7780459} and after fine-tuning it beats Sketch-BERT. For SFT we used learning rate 1e-4 and batch size 4096. The training took between 20h on 256 TPUv5e.
\section{Ablation study}
\label{sec:ablation}
\subsection{Task mixture ablation}
\label{sec:recognition_ablation}
To assess the potential for negative transfer in the mixture, we evaluate the InkFM model's zero-shot performance against PaLIGemma performance when trained on a specific recognition dataset. Table ~\ref{table:reco_ablation} shows that we have comparable performance across the different datasets. To ensure a fair comparison, PaLIGemma SFT models are trained on the same number of examples from each dataset as InkFM model. Table~\ref{table:reco_ablation} demonstrates that LoRA significantly outperforms SFT on smaller datasets like VNOnDB and DeepWriting. These results were used in Section~\ref{sec:recognition_results} to guide the choice between LoRA and SFT for adapting the InkFM model to each dataset.

\subsection{Language information ablation}
\label{sec:language_ablation}
In order to check the generalization to different prompt formulations, we evaluate InkFM model's zero-shot performance on various recognition datasets, both with and without incorporating language information. Table~\ref{table:language_code_abaltion} illustrates that for less common languages, like Vietnamese, the language information decreases the CER more than two times. The explanation to this behavior is that Vietnamese language is part of Latin writing system, that includes hundreds of languages. The language information prompts the model to decode Vietnamese text. In the cases of high-resource languages like English and Chinese, the effects of providing the language information is negligible.

\subsection{Segmentation task formulation}
\label{sec:segmentation_ablation}
As described in Section~\ref{sec:method_note_segmentation}, we utilize two types of tasks for segmentation in the training process – one and many. In the case of ``many'', we expect the model to generate all objects of a given level in one response. In Table~\ref{table:ablation_segmentation_single_multi} we find that for objects of medium level and lowest level there is a decrease in performance (``many'' compared to ``one''). This is explained by the presence of a larger number of smaller objects on the page. However, the performance on larger objects (like text blocks) is comparable in both settings and ``many'' has a benefit of decreased inference time. 

\begin{table}[!h]
  \caption{On the highest level of segmentation, a single inference across all classes is comparable in performance to individual class-specific inferences. There is a significant drop in performance for words due to the increase in sequence length.}
  \label{table:ablation_segmentation_single_multi}
  \small
  \centering
  \begin{tabular}{lcc}
  \toprule
  & \multicolumn{2}{c}{mAP $\uparrow$} \\
   class & one & many \\
    \midrule
    textblock & 77.78 & 77.28 \\
    drawing & 85.69 & 83.79 \\
    list & 92.96 & 92.15 \\
    table & 99.77 & 99.75 \\
    diagram & 95.41 & 96.76 \\
    \midrule
    textline & 84.25 & 83.70 \\
    enclosure & 84.74 & 82.32 \\
    \midrule
    word & 36.55 & 17.32 \\
    box & 86.12 & 80.86 \\
    arrow & 61.88 & 60.12 \\
    \bottomrule
  \end{tabular}
\end{table}
\section{Limitations}
\label{sec:limitations}
Character segmentation is not currently supported as the part of segmentation but may be added in the future.

\section{Conclusion}
\label{sec:conclusion}
We unify segmentation, recognition and classification of handwriting in a single model.  We obtain SoTA quality on handwritten text line segmentation (English), sketch classification and multi-script text recognition on a majority of datasets. Furthermore, the same model can detect visual elements in handwritten notes, including lists, tables, and diagrams. making a unified model a viable solution for the future of handwriting and note taking tasks.

\section*{Ethical Statement}
Online handwriting recognition, segmentation, and classification are well-established technologies.  We do not foresee any negative societal impacts from improving the model's performance of these tasks.

\bibliographystyle{named}
\bibliography{ijcai25}

\end{document}